\documentclass{article}

\usepackage{microtype}
\usepackage{graphicx}
\usepackage{subfigure}
\usepackage{booktabs} 
\usepackage{xspace}
\usepackage{todonotes}
\usepackage{tikz}
\usepackage{pgfplots}

\newcommand{\sparql}{\textsc{Sparql}\xspace}
\newcommand{\arq}{\textsc{Arq}\xspace}
\newcommand{\kbqa}{\textsc{Kbqa}\xspace}
\newcommand{\qald}{\textsc{Qald}\xspace}
\newcommand{\bleu}{\textsc{Bleu}\xspace}
\newcommand{\wqsp}{\textsc{WebQuestionsSP}\xspace}
\newcommand{\lisp}{\textsc{Lisp}\xspace}

\usepackage{hyperref}

\usepackage[accepted]{icml2018}

\icmltitlerunning{Neural Machine Translation for Query Construction and Composition}

\begin{document}

\twocolumn[
\icmltitle{Neural Machine Translation for Query Construction and Composition}



\icmlsetsymbol{equal}{*}

\begin{icmlauthorlist}
\icmlauthor{Tommaso Soru}{aksw}
\icmlauthor{Edgard Marx}{aksw}
\icmlauthor{Andr\'e Valdestilhas}{aksw}
\icmlauthor{Diego Esteves}{sda}
\icmlauthor{Diego Moussallem}{aksw}
\icmlauthor{Gustavo Publio}{aksw}
\end{icmlauthorlist}

\icmlaffiliation{aksw}{AKSW, University of Leipzig, Leipzig, Germany}
\icmlaffiliation{sda}{SDA, Bonn University, Bonn, Germany}

\icmlcorrespondingauthor{Tommaso Soru}{tsoru@informatik.uni-leipzig.de}

\icmlkeywords{Semantic Parsing, Question Answering, Neural Machine Translation}

\vskip 0.3in
]



\printAffiliationsAndNotice{}  

\begin{abstract}
Research on question answering with knowledge base has recently seen an increasing use of deep architectures.
In this extended abstract, we study the application of the neural machine translation paradigm for question parsing.
We employ a sequence-to-sequence model to learn graph patterns in the \sparql graph query language and their compositions.
Instead of inducing the programs through question-answer pairs, we expect a semi-supervised approach, where alignments between questions and queries are built through templates.
We argue that the coverage of language utterances can be expanded using late notable works in natural language generation.
\end{abstract}


\section{Introduction}

Question Answering with Knowledge Base (\kbqa) parses a natural-language question and returns an appropriate answer that can be found in a knowledge base.
Today, one of the most exciting scenarios for question answering is the Web of Data, a fast-growing distributed graph of interlinked knowledge bases which comprises more than 100 billions of edges~\cite{mccrae2018lod}.
Question Answering over Linked Data (\qald) is a subfield of \kbqa aimed at transforming utterances into \sparql queries~\cite{lopez2013evaluating}.
Being a W3C standard, \sparql features a high expressivity~\cite{prud2006sparql} and is by far the most used query language for Linked Data.

Among traditional approaches to \kbqa, \citet{bao2014kbqa} proposed question decomposition and Statistical Machine Translation to translate sub-questions into triple patterns.
The method however relies on entity detection and struggles in recognizing predicates by their contexts (e.g., \emph{play} in a film or a football team).
In the last years, several methods based on neural networks have been devised to tackle the \kbqa problem~\cite{liang2016neural,hao2017end,lukovnikov2017www,sorokin2017end}.
We study the application of the Neural Machine Translation paradigm for question parsing using a sequence-to-sequence model within an architecture dubbed Neural \sparql Machine, previously introduced in~\citet{sorumarx2017}.
Similarly to \citet{liang2016neural}, we employ a sequence-to-sequence model to learn query expressions and their compositions.
Instead of inducing the programs through question-answer pairs, we expect a semi-supervised approach, where alignments between questions and queries are built through templates.
Although query induction can save a considerable amount of supervision effort~\cite{liang2016neural,zhong2017seq2sql}, a pseudo-gold program is not guaranteed to be correct when the same answer can be found with more than one query (e.g., as the \emph{capital} is often the \emph{largest city} of a country, predicates might be confused).
On the contrary, our proposed solution relies on manual annotation and a weakly-supervised expansion of question-query templates.

\begin{figure}[t]
    \centering
    \includegraphics[width=.45\textwidth]{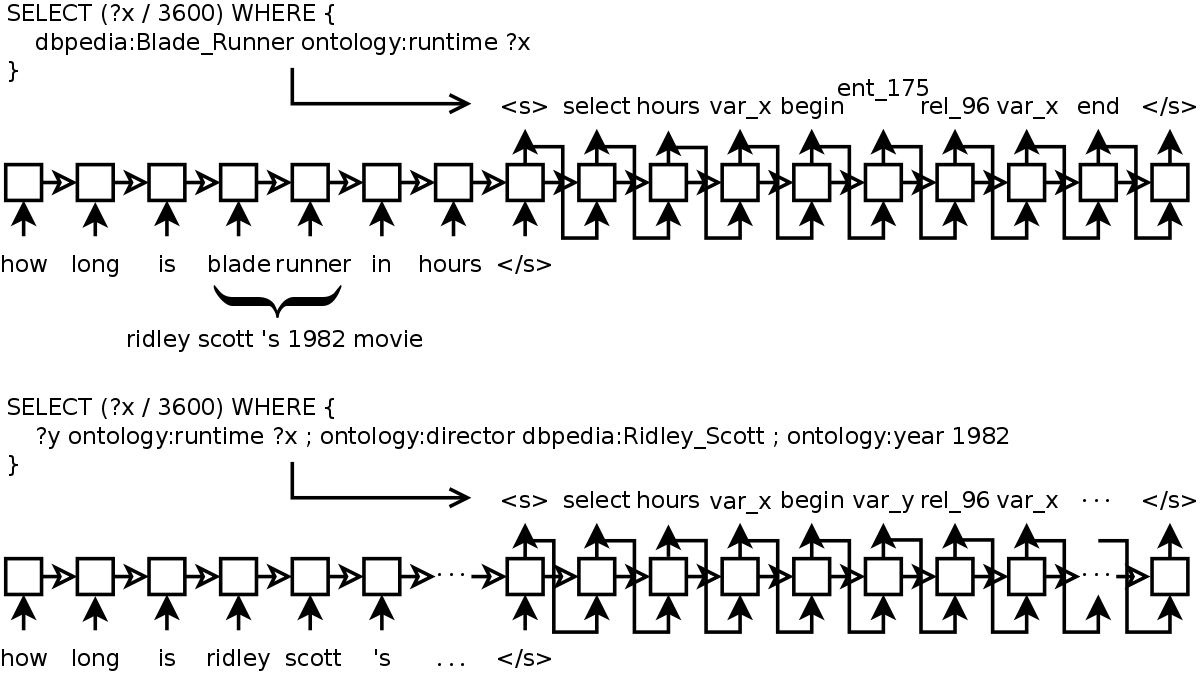}
    \caption{Utterances are translated into \sparql queries encoded as sequences of tokens. Using complex surface forms leads to more graph patterns. We aim at learning these compositions.}
    \label{fig:example}
\end{figure}

\section{Neural SPARQL Machines} \label{sec:nspm}

\begin{table*}[t]
\centering
\caption{Experiments on a DBpedia subset about movies with different \sparql encodings and settings.}
\label{tab:experiments}
\begin{tabular}{@{}llcccc@{}}
\toprule
\textbf{Encoding}         &  \textbf{Description}         & \textbf{Test BLEU} & \textbf{Accuracy} & \textbf{Runtime} & \textbf{Convergence} \\ \midrule
v1           & 1:1 \sparql encoding           & 80.89\%               & 22.33\%             & 1h02:01          & 13,000               \\
v1.1           & Improved consistency & 80.61\%               & 22.33\%             & 1h21:21          & 17,000               \\
v2           & Added templates with $>1$ placeholders  & 89.69\%               & 91.04\%             & 1h59:10          & 22,000               \\
v2.1           & Encoding fix (double spaces removed) & 98.40\%               & 91.05\%             & 1h47:11          & 20,000               \\
v3           & Shortened \sparql sequences   & 99.28\%               & 94.82\%             & 1h12:07          & 25,000               \\
v4           & Added direct entity translations    & 99.29\%               & 93.69\%             & 1h23:00          & 20,000               \\ \bottomrule
\end{tabular}
\end{table*}

Inspired by the \emph{Neural Programmer-Interpreter} pattern by~\cite{reed2015neural}, a Neural \sparql Machine is composed by three modules: a \emph{generator}, a \emph{learner}, and an \emph{interpreter}~\cite{sorumarx2017}.
We define a \emph{query template} as an alignment between a natural language question and its respective \sparql query, with entities replaced by placeholders (e.g., \emph{``where is $<$A$>$ located in?''}).
The generator takes query templates as input and creates the training dataset, which is forwarded to the learner.
The learner takes natural language as input and generates a sequence which encodes a \sparql query.
Here, a recurrent neural network based on (Bidirectional) Long Short-Term Memories~\cite{hochreiter1997long} is employed as a sequence-to-sequence translator (see example in Figure~\ref{fig:example}).
The final structure is then reconstructed by the interpreter through rule-based heuristics.
Note that a sequence can be represented by any \lisp S-expression; therefore, alternatively, sentence dependency trees can be used to encode questions and \arq algebra~\cite{seaborne2010arq} can be used to encode \sparql queries.

Neural \sparql Machines do not rely on entity linking methods, since entities and relations are detected within the query construction phase.
External pre-trained word embeddings help deal with vocabulary mismatch.
Knowledge graph jointly embedded with \sparql operators~\cite{wang2014knowledge} can be utilized in the target space.
A curriculum learning~\cite{bengio2009curriculum} paradigm can learn graph pattern and \sparql operator composition, in a similar fashion of~\citet{liang2016neural}.
We argue that the coverage of language utterances can be expanded using techniques such as Question~\cite{abujabal2017automated,elsahar2018zero,abujabal2018never} and Query Generation~\cite{zafar2018formal} as well as Universal Sentence Encoders~\cite{cer2018universal}.
Another problem is the disambiguation between entities having the same surface forms.
Building on top of the DBtrends approach~\cite{marx2016dbtrends}, we force the number of occurrences of a given entity in the training set to be inversely proportional to the entity ranking.
Following this strategy, we expect the RNN to associate the word \emph{Berlin} with the German capital and not with \emph{Berlin, New Hampshire}.

\section{Experiments and current progress}

We selected the DBpedia Knowledge Base~\cite{lehmann2015dbpedia} as the dataset for our experiments, due to its central importance for the Web of Data.
We built a dataset of 3,108 entities about movies from DBpedia and annotated 20 and 4 question-query templates with one and two placeholders, resp.
Our preliminary results are given in Table~\ref{tab:experiments}.
We experimented with 6 different \sparql encodings, i.e. ways to encode a \sparql query into a sequence of tokens.
At each row of the table, we provide the description of the corresponding changes, each of which persists in the next encodings.
The experiments were carried out on a 64-CPU Ubuntu machine with 512 GB RAM.\footnote{Code available at \url{https://github.com/AKSW/NSpM}.}
We adopted the implementation of \emph{seq2seq} in \emph{TensorFlow} with internal embeddings of 128 dimensions, 2 hidden layers, and a dropout value of 0.2.
All settings were tested on the same set of unseen questions after applying an 80-10-10\% split.

The results confirmed that the \sparql encoding highly influences the learning.
Adding more complex templates (i.e., with more than one placeholder) to the generator input yielded a richer training set and more questions were parsed correctly.
Merging tokens (see queries and their respective sequences in Figure~\ref{fig:example}) helped the machine translation, as the \sparql sequences became shorter.
Adding alignments of entities and their labels to the training set turned out to be beneficial for a faster convergence, as Figure~\ref{fig:bleu} shows.
The most frequent errors were due to entity name collisions and out-of-vocabulary words; both issues can be tackled with the strategies introduced in this work.

\pgfplotsset{scaled x ticks=false}
\begin{figure}
    \centering
    \resizebox {0.4\textwidth} {0.8\height} {
        \begin{tikzpicture}
        \begin{axis}[
        	scale=0.8,
        	ymin=50,
        	ymax=100,
        	legend pos=outer north east,
        ]
        \addplot[mark=., mark size=2, line width=1.5, color=black] table [x=Epoch, y=v1, col sep=tab] {movies-bleu.csv};
        \addplot[mark=., mark size=2, line width=1.5, color=gray] table [x=Epoch, y=v1.1, col sep=tab] {movies-bleu.csv};
        \addplot[mark=., mark size=2, line width=1.5, color=orange] table [x=Epoch, y=v2, col sep=tab] {movies-bleu.csv};
        \addplot[mark=., mark size=2, line width=1.5, color=red] table [x=Epoch, y=v2.1, col sep=tab] {movies-bleu.csv};
        \addplot[mark=., mark size=2, line width=1.5, color=green] table [x=Epoch, y=v3, col sep=tab] {movies-bleu.csv};
        \addplot[mark=., mark size=2, line width=1.5, color=blue] table [x=Epoch, y=v4, col sep=tab] {movies-bleu.csv};
        \legend{v1\\v1.1\\v2\\v2.1\\v3\\v4\\}
        \end{axis}
        \end{tikzpicture}
    }
    \caption{\bleu accuracy against training epochs.}
    \label{fig:bleu}
\end{figure}
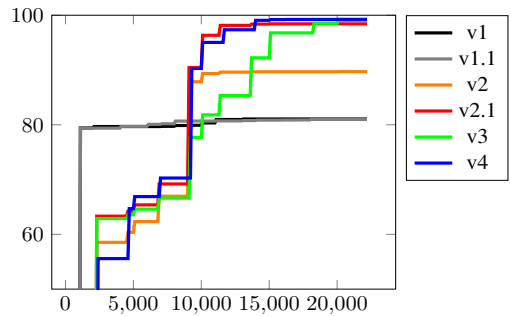

We plan to perform an evaluation on the \wqsp~\cite{yih2016value} and \qald~\cite{unger2014question} benchmarks to compare with the state-of-the-art approaches for \kbqa and \qald, respectively.

\bibliography{biblio}
\bibliographystyle{icml2018}

\end{document}